\documentclass[runningheads]{llncs}
\usepackage{graphicx}
\usepackage{amsmath,amssymb} 
\usepackage{color}

\usepackage{times}
\usepackage{epsfig}
\usepackage{url}
\usepackage{subfigure}

\usepackage[width=122mm,left=12mm,paperwidth=146mm,height=193mm,top=12mm,paperheight=217mm]{geometry}

\newcommand{\etal}{\emph{et al.}}
\newcommand{\eg}{\emph{e.g.,}}
\newcommand{\ie}{\emph{i.e.,}}

\begin{document}
\pagestyle{headings}
\mainmatter
\def\ECCV16SubNumber{359}  

\title{Accelerating the Super-Resolution \\Convolutional Neural Network} 

\titlerunning{Accelerating the Super-Resolution Convolutional Neural Network}

\authorrunning{Chao Dong \etal}

\author{Chao Dong \and Chen Change Loy  \and Xiaoou Tang}
\institute{Department of Information Engineering, The Chinese University of Hong Kong\\
	\email{ \{dc012,ccloy,xtang\}@ie.cuhk.edu.hk}
}

\maketitle

\begin{abstract}

As a successful deep model applied in image super-resolution (SR), the Super-Resolution Convolutional Neural Network (SRCNN)~\cite{Dong2014,Dong2015} has demonstrated superior performance to the previous hand-crafted models either in speed and restoration quality. However, the high computational cost still hinders it from practical usage that demands real-time performance (24 fps). In this paper, we aim at accelerating the current SRCNN, and propose a compact hourglass-shape CNN structure for faster and better SR. We re-design the SRCNN structure mainly in three aspects. First, we introduce a deconvolution layer at the end of the network, then the mapping is learned directly from the original low-resolution image (without interpolation) to the high-resolution one. Second, we reformulate the mapping layer by shrinking the input feature dimension before mapping and expanding back afterwards. Third, we adopt smaller filter sizes but more mapping layers. The proposed model achieves a speed up of more than 40 times with even superior restoration quality. Further, we present the parameter settings that can achieve real-time performance on a generic CPU while still maintaining good performance. A corresponding transfer strategy is also proposed for fast training and testing across different upscaling factors.
\end{abstract}

\section{Introduction}
Single image super-resolution (SR) aims at recovering a
high-resolution (HR) image from a given low-resolution (LR)
one. Recent SR algorithms are mostly
learning-based (or patch-based) methods~\cite{Dong2014,Dong2015,yang2013fast,Timofte2013,Timofte2014,Cui2014,Schulter2015,Wang2015}
that learn a mapping between the LR and HR image spaces.
Among them, the Super-Resolution Convolutional Neural Network (SRCNN)~\cite{Dong2014,Dong2015}  has drawn considerable attention due to its simple network structure and excellent restoration quality. Though SRCNN is already faster than most previous learning-based methods, the processing speed on large images
is still unsatisfactory. For example, to upsample an $240\times 240$ image by a factor of 3, the speed of the original SRCNN~\cite{Dong2014} is about 1.32 fps, which is far from real-time (24 fps).
To approach real-time, we should accelerate SRCNN for at least 17 times while keeping the previous performance.
This sounds implausible at the first glance, as accelerating by simply reducing the parameters will severely impact the performance.
However, when we delve into the network structure, we find \textit{two inherent limitations} that restrict its running speed.

\begin{figure}[t]
\centering
  \includegraphics[width=0.75\linewidth]{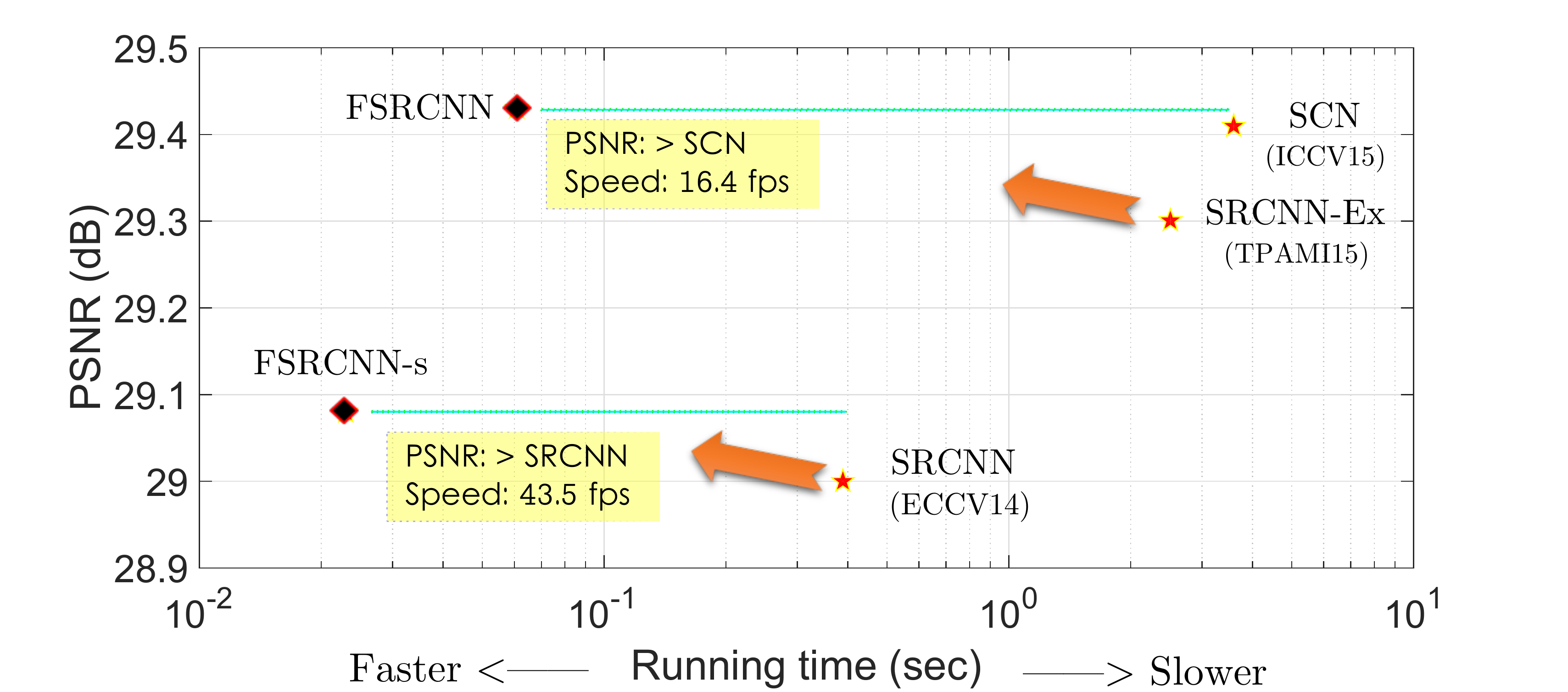}
\caption{The proposed FSRCNN networks achieve better super-resolution quality than existing methods, and are tens of times faster. Especially, the FSRCNN-s can run in real-time ($>24$ fps) on a generic CPU. The chart is based on the Set14~\cite{Zeyde2012} results summarized in Tables~\ref{results91} and~\ref{results}. }
  \label{fig:runtime}
\end{figure}

First, as a pre-processing step, the original LR image needs to
be upsampled to the desired size using bicubic interpolation to form the input. Thus the computation complexity of SRCNN grows quadratically with the spatial size of the HR image (not the original LR image). For the upscaling factor $n$, the computational cost of convolution with the interpolated LR image will be $n^2$ times of that for the original LR one. This is also the restriction for most learning-based SR methods~\cite{Yang2010a,yang2013fast,Timofte2013,Timofte2014,Schulter2015,Wang2015}. If the network was learned directly from the original LR image, the acceleration would be significant, \ie~about $n^2$ times faster.

The second restriction lies on the costly non-linear mapping step. In SRCNN, input image patches are projected on a high-dimensional LR feature space, then followed by a complex mapping to another high-dimensional HR feature space. Dong~\etal~\cite{Dong2015} show that the mapping accuracy can be substantially improved by adopting a wider mapping layer, but at the cost of the running time. For example, the large SRCNN (SRCNN-Ex)~\cite{Dong2015} has 57,184 parameters, which are six times larger than that for SRCNN (8,032 parameters). Then the question is how to shrink the network scale while still keeping the previous accuracy.

According to the above observations, we investigate a more concise and efficient network structure for fast and accurate image SR.
\textit{To solve the first problem}, we adopt a deconvolution layer to replace the bicubic interpolation. To further ease the computational burden, we place the deconvolution layer\footnote{We follow~\cite{Zeiler2014} to adopt the terminology `deconvolution'. We note that it carries very different meaning in classic image processing, see~\cite{xu2014deep}.} at the end of the network, then the computational complexity is only proportional to the spatial size of the original LR image. It is worth noting that the deconvolution layer is not equal to a simple substitute of the conventional interpolation kernel like in FCN~\cite{Long2015}, or `unpooling+convolution' like~\cite{dosovitskiy2015learning}. Instead, it consists of diverse automatically learned upsampling kernels (see Figure~\ref{fig:filters}) that work jointly to generate the final HR output, and replacing these deconvolution filters with uniform interpolation kernels will result in a drastic PSNR drop (\eg~at least 0.9 dB on the Set5 dataset~\cite{Bevilacqua2012} for $\times 3$).

\textit{For the second problem}, we add a shrinking and an expanding layer at the beginning and the end of the mapping layer separately to restrict mapping in a low-dimensional feature space. Furthermore, we decompose a single wide mapping layer into several layers with a fixed filter size $3\times 3$. The overall shape of the new structure looks like an \textit{hourglass}, which is symmetrical on the whole, thick at the ends and thin in the middle. Experiments show that the proposed model, named as Fast Super-Resolution Convolutional Neural Networks (FSRCNN)~\footnote{The implementation is available on the project page \url{http://mmlab.ie.cuhk.edu.hk/projects/FSRCNN.html}.}, achieves a speed-up of more than $40\times$ with even superior performance than the SRCNN-Ex. In this work, we also present a small FSRCNN network (FSRCNN-s) that achieves similar restoration quality as SRCNN, but is $17.36$ times faster and can run in real time (24 fps) with a generic CPU.
As shown in Figure~\ref{fig:runtime}, the FSRCNN networks are much faster than contemporary SR models yet achieving superior performance.

Apart from the notable improvement in speed, the FSRCNN also has another appealing property that could facilitate fast training and testing across different upscaling factors. Specifically, in FSRCNN, all convolution layers (except the deconvolution layer) can be shared by networks of different upscaling factors. During training, with a well-trained network, we only need to fine-tune the deconvolution layer for another upscaling factor with almost no loss of mapping accuracy. During testing, we only need to do convolution operations once, and upsample an image to different scales using the corresponding deconvolution layer.

Our contributions are three-fold: 1) We formulate a compact hourglass-shape CNN structure for fast image super-resolution. With the collaboration of a set of deconvolution filters, the network can learn an end-to-end mapping between the original LR and HR images with no pre-processing.
2) The proposed model achieves a speed up of at least $40\times$ than the SRCNN-Ex~\cite{Dong2015} while still keeping its exceptional performance. One of its small-size version can run in real-time ($>$24 fps) on a generic CPU with better restoration quality than SRCNN~\cite{Dong2014}.
3) We transfer the convolution layers of the proposed networks for fast training and testing across different upscaling factors, with no loss of restoration quality.

\section{Related Work}

\noindent
\textbf{Deep learning for SR:}
Recently, the deep learning techniques have been successfully applied on SR. The pioneer work is termed as the Super-Resolution Convolutional Neural Network (SRCNN) proposed by Dong~\etal~\cite{Dong2014,Dong2015}. Motivated by SRCNN, some problems such as face hallucination~\cite{Zhu2016} and depth map super-resolution~\cite{Hui2016} have achieved state-of-the-art results. Deeper structures have also been explored in~\cite{Kim2015a} and~\cite{Kim2015}. Different from the conventional learning-based methods, SRCNN directly learns an end-to-end mapping between LR and HR images, leading to a fast and accurate inference. The inherent relationship between SRCNN and the sparse-coding-based methods ensures its good performance.   Based on the same assumption, Wang~\etal~\cite{Wang2015} further replace the mapping layer by a set of sparse coding sub-networks and propose a sparse coding based network (SCN). With the domain expertise of the conventional sparse-coding-based method, it outperforms SRCNN with a smaller model size. However, as it strictly mimics the sparse-coding solver, it is very hard to shrink the sparse coding sub-network with no loss of mapping accuracy. Furthermore, all these networks~\cite{Wang2015,Kim2015a,Kim2015} need to process the bicubic-upscaled LR images.
The proposed FSRCNN does not only perform on the original LR image, but also contains a simpler but more efficient mapping layer. Furthermore, the previous methods have to train a totally different network for a specific upscaling factor, while the FSRCNN only requires a different deconvolution layer. This also provides us a faster way to upscale an image to several different sizes.

\noindent
\textbf{CNNs acceleration:} A number of studies have investigated the acceleration of CNN. Denton~\etal~\cite{Denton2014} first investigate the redundancy within the CNNs designed for object detection. Then Zhang~\etal~\cite{Zhang2015} make attempts to accelerate very deep CNNs for image classfication. They also take the non-linear units into account and reduce the accumulated error by asymmetric reconstruction.
Our model also aims at accelerating CNNs but in a different manner. First, they focus on approximating the existing well-trained models, while we reformulate the previous model and achieves better performance. Second, the above methods are all designed for high-level vision problems (\eg~image classification and object detection), while ours are for the low-level vision task. As the deep models for SR contain no fully-connected layers, the approximation of convolution filters will severely impact the performance. 

\section{Fast Super-Resolution by CNN}

We first briefly describe the network structure of SRCNN~\cite{Dong2014,Dong2015}, and then we detail how we reformulate the network layer by layer. The differences between FSRCNN and SRCNN are presented at the end of this section.

\subsection{SRCNN}
\begin{figure*}[t]
\centering
  \includegraphics[width=1\linewidth]{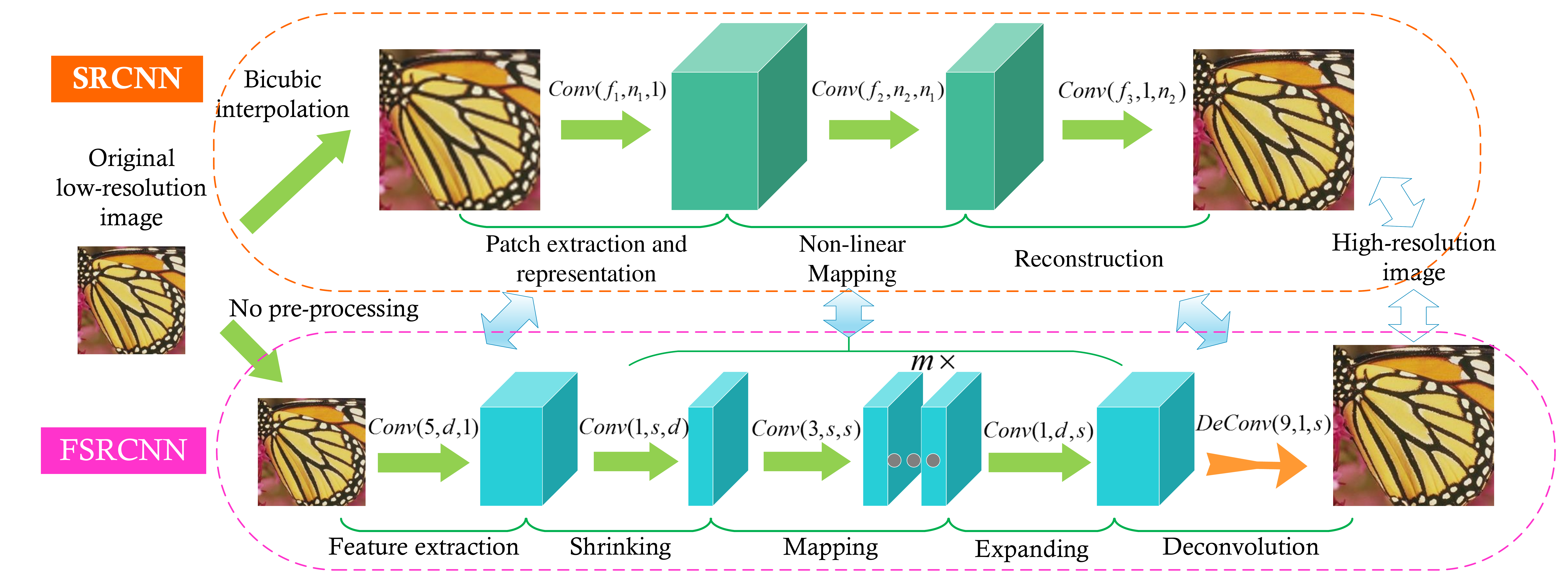}
  \caption{This figure shows the network structures of the SRCNN and FSRCNN. The proposed FSRCNN is different from SRCNN mainly in three aspects. First, FSRCNN adopts the original low-resolution image as input without bicubic interpolation. A deconvolution layer is introduced at the end of the network to perform upsampling. Second, The non-linear mapping step in SRCNN is replaced by three steps in FSRCNN, namely the shrinking, mapping, and expanding step. Third, FSRCNN adopts smaller filter sizes and a deeper network structure. These improvements provide FSRCNN with better performance but lower computational cost than SRCNN.}
  \label{fig:structure}
\end{figure*}

SRCNN aims at learning an end-to-end mapping function $F$ between the bicubic-interpolated LR image $Y$ and the HR image $X$. The network contains all convolution layers, thus the size of the output is the same as that of the input image.  As depicted in Figure~\ref{fig:structure}, the overall structure consists of three parts that are analogous to the main steps of the sparse-coding-based methods~\cite{Yang2010a}. The patch extraction and representation part refers to the first layer, which extracts patches from the input and represents each patch as a high-dimensional feature vector. The non-linear mapping part refers to the middle layer, which maps the feature vectors non-linearly to another set of feature vectors, or namely HR features. Then the last reconstruction part aggregates these features to form the final output image.

The computation complexity of the network can be calculated as follows,
\begin{equation}
\label{eqn:computationSRCNN}
O\{(f_1^2 n_1 + n_1  f_2^2  n_2 + n_2 f_3^2) S_{HR}\},
\end{equation}
where $\{f_{i}\}_{i=1}^3$ and $\{n_{i}\}_{i=1}^3$ are the filter size and filter number of the three layers, respectively. $S_{HR}$ is the size of the HR image. We observe that the complexity is proportional to the size of the HR image, and the middle layer contributes most to the network parameters. In the next section, we present the FSRCNN by giving special attention to these two facets.

\subsection{FSRCNN}
As shown in Figure~\ref{fig:structure}, FSRCNN can be decomposed into five parts -- feature extraction, shrinking, mapping, expanding and deconvolution. The first four parts are convolution layers, while the last one is a deconvolution layer. For better understanding, we denote a convolution layer as $Conv(f_i,n_i,c_i)$, and a deconvolution layer as $DeConv(f_i,n_i,c_i)$, where the variables $f_i,n_i,c_i$ represent the filter size, the number of filters and the number of channels, respectively.

As the whole network contains tens of variables (\ie~$\{f_i,n_i,c_i\}_{i=1}^6$), it is impossible for us to investigate each of them. Thus we assign a reasonable value to the insensitive variables in advance, and leave the sensitive variables unset. We call a variable sensitive when a slight change of the variable could significantly influence the performance. These sensitive variables always represent some important influential factors in SR, which will be shown in the following descriptions.

\noindent
\textbf{Feature extraction:} This part is similar to the first part of SRCNN, but different on the input image. FSRCNN performs feature extraction on the original LR image without interpolation. To distinguish from SRCNN, we denote the small LR input as $Y_s$. By doing convolution with the first set of filters, each patch of the input (1-pixel overlapping) is represented as a high-dimensional feature vector.

We refer to SRCNN on the choice of parameters -- $f_1,n_1,c_1$. In SRCNN, the filter size of the first layer is set to be 9. Note that these filters are performed on the upscaled image $Y$. As most pixels in $Y$ are interpolated from $Y_s$, a $5\times 5$ patch in $Y_s$ could cover almost all information of a $9\times 9$ patch in $Y$. Therefore, we can adopt a smaller filter size $f_1=5$ with little information loss. For the number of channels, we follow SRCNN to set $c_1=1$. Then we only need to determine the filter number $n_1$. From another perspective, $n_1$ can be regarded as the number of LR feature dimension, denoted as $d$ -- the first sensitive variable. Finally, the first layer can be represented as $Conv(5,d,1)$.

\noindent
\textbf{Shrinking:} In SRCNN, the mapping step generally follows the feature extraction step, then the high-dimensional LR features are mapped directly to the HR feature space. However, as the LR feature dimension $d$ is usually very large, the computation complexity of the mapping step is pretty high. This phenomenon is also observed in some deep models for high-level vision tasks.
Authors in \cite{Lin2014} apply $1\times 1$ layers to save the computational cost.

With the same consideration, we add a shrinking layer after the feature extraction layer to reduce the LR feature dimension $d$. We fix the filter size to be $f_2=1$, then the filters perform like a linear combination within the LR features. By adopting a smaller filter number $n_2=s<<d$, the LR feature dimension is reduced from $d$ to $s$. Here $s$ is the second sensitive variable that determines the level of shrinking, and the second layer can be represented as $Conv(1,s,d)$.
This strategy greatly reduces the number of parameters (detailed computation in Section~\ref{sec:parameters}).

\noindent
\textbf{Non-linear mapping:} The non-linear mapping step is the most important part that affects the SR performance, and the most influencing factors are the width (\ie~the number of filters in a layer) and depth (\ie~the number of layers) of the mapping layer. As indicated in SRCNN~\cite{Dong2015}, a $5\times 5$ layer achieves much better results than a $1\times 1$ layer. But they are lack of experiments on very deep networks.
The above experiences help us to formulate a more efficient mapping layer for FSRCNN. First, as a trade-off between the performance and network scale, we adopt a medium filter size $f_3=3$. Then, to maintain the same good performance as SRCNN, we use multiple $3\times 3$ layers to replace a single wide one. The number of mapping layers is another sensitive variable (denoted as $m$), which determines both the mapping accuracy and complexity. To be consistent, all mapping layers contain the same number of filters $n_3=s$. Then the non-linear mapping part can be represented as $m\times Conv(3, s, s)$.

\noindent
\textbf{Expanding:} The expanding layer acts like an inverse process of the shrinking layer. The shrinking operation reduces the number of LR feature dimension for the sake of the computational efficiency. However, if we generate the HR image directly from these low-dimensional features, the final restoration quality will be poor. Therefore, we add an expanding layer after the mapping part to expand the HR feature dimension. To maintain consistency with the shrinking layer, we also adopt $1\times 1$ filters, the number of which is the same as that for the LR feature extraction layer. As opposed to the shrinking layer $Conv(1,s,d)$, the expanding layer is $Conv(1,d,s)$. Experiments show that without the expanding layer, the performance decreases up to 0.3 dB on the Set5 test set~\cite{Bevilacqua2012}.

\begin{figure}[t]
\centering
  \includegraphics[width=0.8\linewidth]{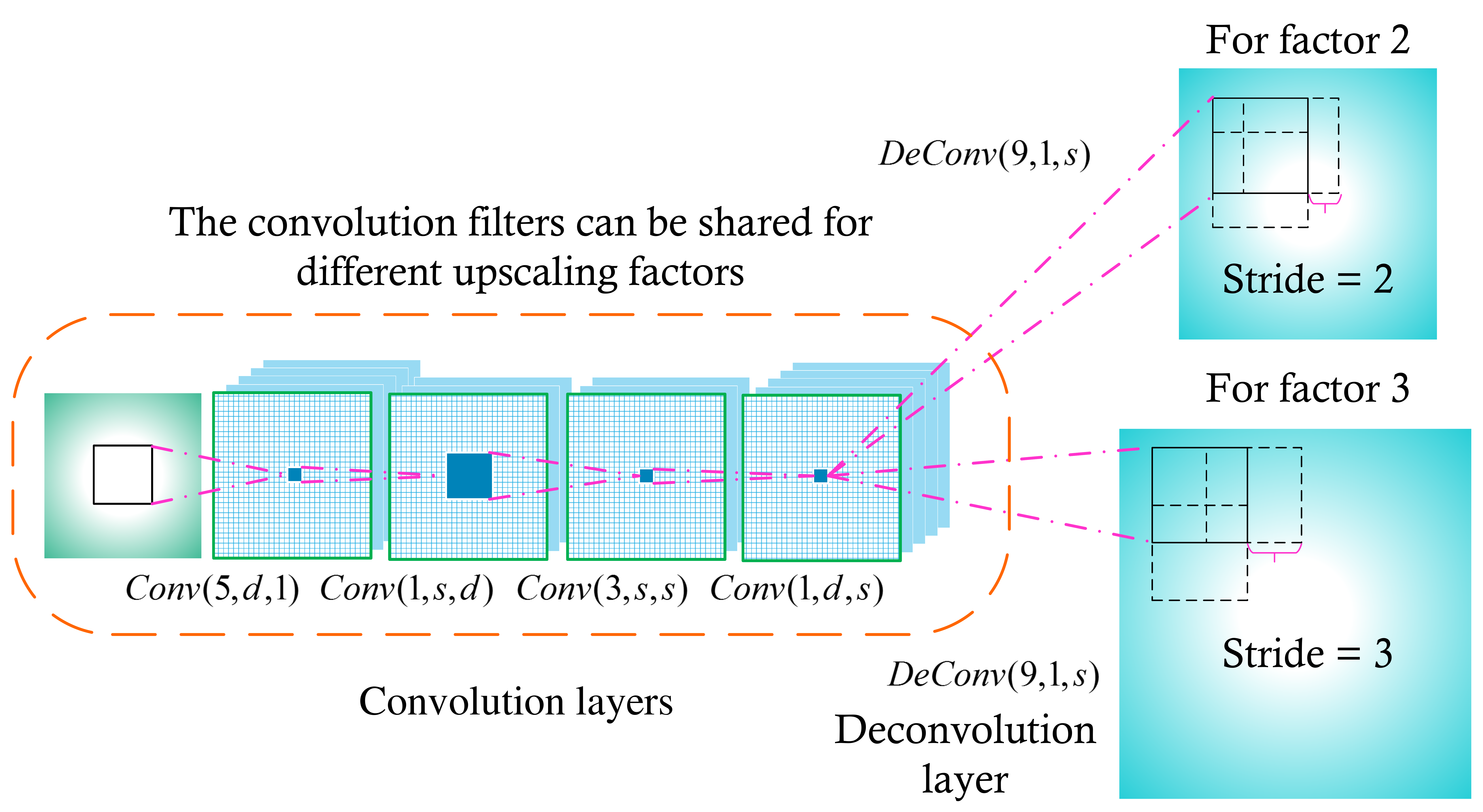}
\caption{The FSRCNN consists of convolution layers and a deconvolution layer. The convolution layers can be shared for different upscaling factors. A specific deconvolution layer is trained for different upscaling factors.}
  \label{fig:deconvolution}
\end{figure}

\noindent
\textbf{Deconvolution:} The last part is a deconvolution layer, which upsamples and aggregates the previous features with a set of deconvolution filters. The deconvolution can be regarded as an inverse operation of the convolution. For convolution, the filter is convolved with the image with a stride $k$, and the output is $1/k$ times of the input. Contrarily, if we exchange the position of the input and output, the output will be $k$ times of the input, as depicted in Figure~\ref{fig:deconvolution}. We take advantage of this property to set the stride $k=n$, which is the desired upscaling factor. Then the output is directly the reconstructed HR image.

\begin{figure}[t]
\centering
  \includegraphics[width=0.7\linewidth]{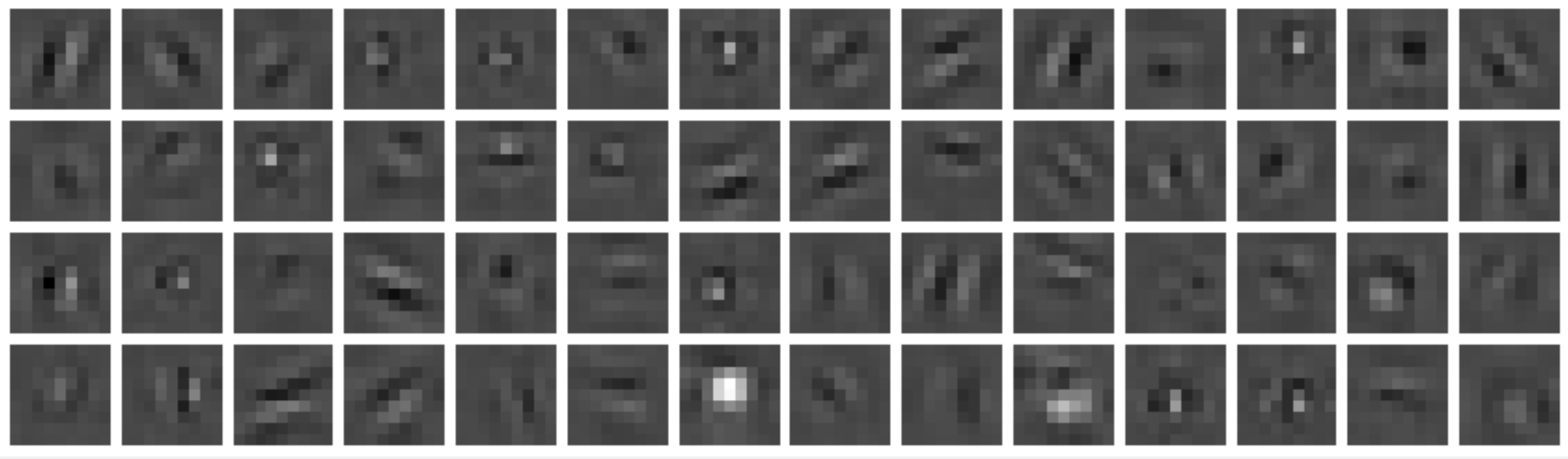}
\caption{The learned deconvolution layer (56 channels) for the upscaling factor 3.}
  \label{fig:filters}
\end{figure}

When we determine the filter size of the deconvolution filters, we can look at the network from another perspective. Interestingly, the reversed network is like a downscaling operator that accepts an HR image and outputs the LR one. Then the deconvolution layer becomes a convolution layer with a stride $n$. As it extracts features from the HR image, we should adopt $9\times 9$ filters that are consistent with the first layer of SRCNN. Similarly, if we reverse back, the deconvolution filters should also have a spatial size $f_5=9$. Experiments also demonstrate this assumption. Figure~\ref{fig:filters} shows the learned deconvolution filters, the patterns of which are very similar to that of the first-layer filters in SRCNN.
Lastly, we can represent the deconvolution layer as $DeConv(9,1,d)$.

Different from inserting traditional interpolation kernels (\eg~bicubic or bilinear) in-network~\cite{Long2015} or having `unpooling+convolution'~\cite{dosovitskiy2015learning}, the deconvolution layer learns a set of upsampling kernel for the input feature maps. As shown in Figure~\ref{fig:filters}, these kernels are diverse and meaningful. If we force these kernels to be identical, the parameters will be used inefficiently (equal to sum up the input feature maps as one), and the performance will drop at least 0.9 dB on the Set5.

\noindent
\textbf{PReLU:} For the activation function after each convolution layer, we suggest the use of the Parametric Rectified Linear Unit (PReLU)~\cite{He2015} instead of the commonly-used Rectified Linear Unit (ReLU). They are different on the coefficient of the negative part. For ReLU and PReLU, we can define a general activation function as $f(x_i)=max(x_i,0)+a_imin(0,x_i)$,
where $x_i$ is the input signal of the activation $f$ on the $i$-th channel, and $a_i$ is the coefficient of the negative part. The parameter $a_i$ is fixed to be zero for ReLU, but is learnable for PReLU. We choose PReLU mainly to avoid the ``dead features''~\cite{Zeiler2014} caused by zero gradients in ReLU. Then we can make full use of all parameters to test the maximum capacity of different network designs. Experiments show that the performance of the PReLU-activated networks is more stable, and can be seen as the up-bound of that for the ReLU-activated networks.

\noindent
\textbf{Overall structure:} We can connect the above five parts to form a complete FSRCNN network as $Conv(5,d,1)-PReLU-Conv(1,s,d)-PReLU-m\times Conv(3,s,s)-PReLU-Conv(1,d,s)-PReLU-DeConv(9,1,d)$.
On the whole, there are three sensitive variables (\ie~the LR feature dimension $d$, the number of shrinking filters $s$, and the mapping depth $m$) governing the performance and speed. For simplicity, we represent a FSRCNN network as $FSRCNN(d,s,m)$. The computational complexity can be calculated as
\begin{equation}
\begin{array}{rcl}
\label{eqn:computationFSRCNN}
O\{(25d + sd + 9ms^2 + ds+ 81d)S_{LR}\} = O\{(9ms^2 + 2sd + 106d)S_{LR}\}.
\end{array}
\end{equation}
We exclude the parameters of PReLU , which introduce negligible computational cost.
Interestingly, the new structure looks like an \textit{hourglass}, which is symmetrical on the whole, thick at the ends, and thin in the middle. The three sensitive variables are just the controlling parameters for the appearance of the hourglass. Experiments show that this hourglass design is very effective for image super-resolution.

\noindent
\textbf{Cost function:} Following SRCNN, we adopt the mean square error (MSE) as the cost function. The optimization objective is represented as
\begin{equation}
\label{eqn:loss}
\min_{\theta} \frac{1}{n}\sum\nolimits_{i=1}^n||F(Y_s^{i} ;\theta) - X^i||_2^2,
\end{equation}
where $Y_s^i$ and $X^i$ are the $i$-th LR and HR sub-image pair in the training data, and $F(Y_s^{i} ; \theta)$ is the network output for $Y_s^i$ with parameters $\theta$. All parameters are optimized using stochastic gradient descent with the standard backpropagation.

\subsection{Differences against SRCNN: From SRCNN to FSRCNN}
\label{sec:parameters}
To better understand how we accelerate SRCNN, we transform the SRCNN-Ex to another FSRCNN (56,12,4) within three steps, and show how much acceleration and PSNR gain are obtained by each step. We use a representative upscaling factor $n=3$. The network configurations of SRCNN, FSRCNN and the two transition states are shown in Table~\ref{tab:transition}. We also show their performance (average PSNR on Set5) trained on the 91-image dataset~\cite{Yang2010a}.

\begin{table}[t]\scriptsize
\caption{The transitions from SRCNN to FSRCNN.}\label{tab:transition}
\begin{center}
\begin{tabular}{|c|c|c|c|c|}
\hline
 &  SRCNN-Ex &  Transition State 1 &  Transition State 2 &  FSRCNN (56,12,4)\\
\hline
First part & Conv(9,64,1) & Conv(9,64,1) & Conv(9,64,1) & \textbf{Conv(5,56,1)} \\
\hline
           &              &              & \textbf{Conv(1,12,64)-} & \textbf{Conv(1,12,56)-} \\
Mid part   & Conv(5,32,64)& Conv(5,32,64)& \textbf{4Conv(3,12,12)} & 4Conv(3,12,12) \\
           &              &              & \textbf{-Conv(1,64,12)} & \textbf{-Conv(1,56,12)} \\
\hline
Last part  & Conv(5,1,32) & \textbf{DeConv(9,1,32)} & \textbf{DeConv(9,1,64)} & \textbf{DeConv(9,1,56)} \\
\hline
Input size  & $S_{HR}$ & $S_{LR}$ & $S_{LR}$ & $S_{LR}$ \\
\hline
Parameters  & 57184 & 58976 & 17088 & 12464 \\
\hline\hline
Speedup &1$\times$ &8.7$\times$ &30.1$\times$ &41.3$\times$ \\
\hline
PSNR (Set5) & 32.83 dB & 32.95 dB & 33.01 dB & 33.06 dB \\
\hline
\end{tabular}
\end{center}
\end{table}

First, we replace the last convolution layer of SRCNN-Ex with a deconvolution layer, then the whole network will perform on the original LR image and the computation complexity is proportional to $S_{LR}$ instead of $S_{HR}$. This step will enlarge the network scale but achieve a speedup of $8.7\times$ (\ie~$57184/58976\times3^2$). As the learned deconvolution kernels are better than a single bicubic kernel, the performance increases roughly by 0.12 dB. Second, the single mapping layer is replaced with the combination of a shrinking layer, 4 mapping layers and an expanding layer. Overall, there are 5 more layers, but the parameters are decreased from 58,976 to 17,088. Also, the acceleration after this step is the most prominent -- $30.1\times$. It is widely observed that depth is the key factor that affects the performance. Here, we use four ``narrow'' layers to replace a single ``wide'' layer, thus achieving better results (33.01 dB) with much less parameters. Lastly, we adopt smaller filter sizes and less filters (\eg~from $Conv(9,64,1)$ to $Conv(5,56,1)$), and obtain a final speedup of $ 41.3\times$. As we remove some redundant parameters, the network is trained more efficiently and achieves another 0.05 dB improvement.

It is worth noting that this acceleration is NOT at the cost of performance degradation. Contrarily, the FSRCNN (56,12,4) outperforms SRCNN-Ex by a large margin (\eg~0.23dB on the Set5 dataset). The main reasons of high performance have been presented in the above analysis. This is the main difference between our method and other CNN acceleration works~\cite{Denton2014,Zhang2015}. Nevertheless, with the guarantee of good performance, it is easier to cooperate with other acceleration methods to get a faster model.

\subsection{SR for Different Upscaling Factors}
\label{subsec:across_factor}

Another advantage of FSRCNN over the previous learning-based methods is that FSRCNN could achieve fast training and testing across different upscaling factors.
Specifically, we find that all convolution layers on the whole act like a complex feature extractor of the LR image, and only the last deconvolution layer contains the information of the upscaling factor. This is also proved by experiments, of which the convolution filters are almost the same for different upscaling factors\footnote{Note that in SRCNN and SCN, the convolution filters differ a lot for different upscaling factors.}. With this property, we can transfer the convolution filters for fast training and testing.

In practice, we train a model for an upscaling factor in advance. Then during training, we only fine-tune the deconvolution layer for another upscaling factor and leave the convolution layers unchanged. The fine-tuning is fast, and the performance is as good as training from scratch (see Section~\ref{sec:transfer}). During testing, we perform the convolution operations once, and upsample an image to different sizes with the corresponding deconvolution layer. If we need to apply several upscaling factors simultaneously, this property can lead to much faster testing (as illustrated in Figure~\ref{fig:deconvolution}).

\section{Experiments}

\subsection{Implementation Details}
\noindent
\textbf{Training dataset.} The 91-image dataset is widely used as the training set in learning-based SR methods~\cite{Yang2010a,Timofte2014,Dong2014}.
As deep models generally benefit from big data, studies have found that 91 images are not enough to push a deep model to the best performance. Yang~\etal~\cite{Yang2014} and Schulter~\etal~\cite{Schulter2015} use the BSD500 dataset~\cite{martin2001database}. However, images in the BSD500 are in JPEG format, which are not optimal for the SR task.
Therefore, we contribute a new General-100 dataset that contains 100 bmp-format images (with no compression)\footnote{We follow \cite{Huang2015} to introduce only 100 images in a new super-resolution dataset. A larger dataset with more training images will be released on the project page.}. The size of the newly introduced 100 images ranges from $710\times 704$ (large) to $131\times 112$ (small). They are all of good quality with clear edges but fewer smooth regions (\eg~sky and ocean), thus are very suitable for the SR training. In the following experiments, apart from using the 91-image dataset for training, we will also evaluate the applicability of the joint set of the General-100 dataset and the 91-image dataset to train our networks.
To make full use of the dataset, we also adopt data augmentation as in \cite{Wang2015}. We augment the data in two ways. 1) Scaling: each image is downscaled with the factor 0.9, 0,8, 0.7 and 0.6. 2) Rotation: each image is rotated with the degree of 90, 180 and 270. Then we will have $5\times 4-1=19$ times more images for training.

\noindent
\textbf{Test and validation dataset.} Following SRCNN and SCN, we use the Set5~\cite{Bevilacqua2012}, Set14 \cite{Zeyde2012} and BSD200~\cite{martin2001database} dataset for testing. Another 20 images from the validation set of the BSD500 dataset are selected for validation.

\noindent
\textbf{Training samples.} To prepare the training data, we first downsample the original training images by the desired scaling factor $n$ to form the LR images. Then we crop the LR training images into a set of $f_{sub}\times f_{sub}$-pixel sub-images with a stride $k$. The corresponding HR sub-images (with size $(nf_{sub})^2$) are also cropped from the ground truth images. These LR/HR sub-image pairs are the primary training data.

For the issue of padding, we empirically find that padding the input or output maps does little effect on the final performance. Thus we adopt zero padding in all layers according to the filter size. In this way, there is no need to change the sub-image size for different network designs.
Another issue affecting the sub-image size is the deconvolution layer. As we train our models with the \textit{Caffe} package~\cite{Jia2014}, its deconvolution filters will generate the output with size $(nf_{sub}-n+1)^2$ instead of $(nf_{sub})^2$. So we also crop $(n-1)$-pixel borders on the HR sub-images. Finally, for $\times$2, $\times$3 and $\times$4, we set the size of LR/HR sub-images to be $10^2/19^2$, $7^2/19^2$ and $6^2/21^2$, respectively.

\noindent
\textbf{Training strategy.}
For fair comparison with the state-of-the-arts (Sec.~\ref{subsec:sota}), we adopt the 91-image dataset for training. In addition, we also explore a two-step training strategy. First, we train a network from scratch with the 91-image dataset. Then, when the training is saturated, we add the General-100 dataset for fine-tuning. With this strategy, the training converges much earlier than training with the two datasets from the beginning.

When training with the 91-image dataset, the learning rate of the convolution layers is set to be $10^{-3}$ and that of the deconvolution layer is $10^{-4}$. Then during fine-tuning, the learning rate of all layers is reduced by half. For initialization, the weights of the convolution filters are initialized with the method designed for PReLU in \cite{He2015}. As we do not have activation functions at the end, the deconvolution filters are initialized by the same way as in SRCNN (\ie~drawing randomly from a Gaussian distribution with zero mean and standard deviation 0.001).

\subsection{Investigation of Different Settings}
\label{sec:Investigation}
To test the property of the FSRCNN structure, we design a set of controlling experiments with different values of the three sensitive variables -- the LR feature dimension $d$, the number of shrinking filters $s$, and the mapping depth $m$. Specifically, we choose $d=48,56$, $s=12,16$ and $m=2,3,4$, thus we conduct a total of $2\times 2\times 3=12$ experiments with different combinations.

\begin{figure}[t]\footnotesize
\centering
  \includegraphics[width=\linewidth]{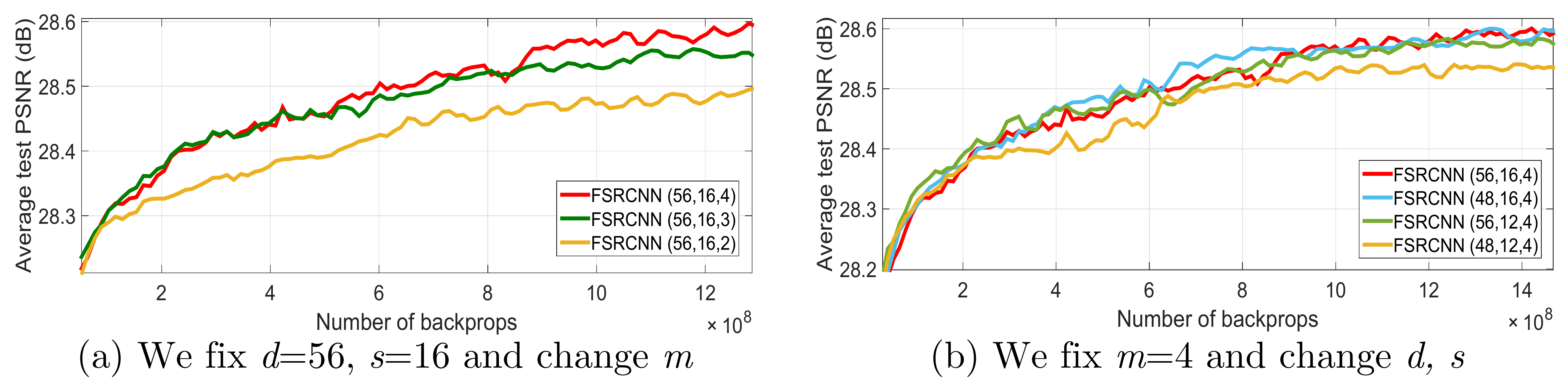}
  \caption{Convergence curves of different network designs.}
  \label{fig:com}
\end{figure}

The average PSNR values on the Set5 dataset of these experiments are shown in Table~\ref{tab:settings}. We analyze the results in two directions,~\ie~horizontally and vertically in the table. First, we fix $d,s$ and examine the influence of $m$. Obviously, $m=4$ leads to better results than $m=2$ and $m=3$. This trend can also be observed from the convergence curves shown in Figure~\ref{fig:com}(a).
Second, we fix $m$ and examine the influence of $d$ and $s$. In general, a better result usually requires more parameters (\eg~a larger $d$ or $s$), but more parameters do not always guarantee a better result. This trend is also reflected in Figure~\ref{fig:com}(b), where we see the three largest networks converge together. From all the results, we find the best trade-off between performance and parameters -- FSRCNN (56,12,4), which achieves one of the highest results with a moderate number of parameters.

It is worth noticing that the smallest network FSRCNN (48,12,2) achieves an average PSNR of 32.87 dB, which is already higher than that of SRCNN-Ex (32.75 dB) reported in~\cite{Dong2015}. The FSRCNN (48,12,2) contains only 8,832 parameters, then the acceleration compared with SRCNN-Ex is $57184/8832\times 9=58.3$ times.

\begin{table}[t]
\caption{The comparison of PSNR (Set5) and parameters of different settings.}\label{tab:settings}
\begin{center}
\begin{tabular}{|c|c|c|c|}
\hline
  Settings &  $m=2$ & $m=3$ & $m=4$ \\

\hline
$d=48,s=12$ & 32.87 (8832) & 32.88 (10128) & 33.08 (11424)\\
\hline
$d=56,s=12$ & 33.00 (9872) & 32.97 (11168) & 33.16 (12464)\\
\hline
$d=48,s=16$ & 32.95 (11232) & 33.10 (13536) & 33.18 (15840)\\
\hline
$d=56,s=16$ & 33.01 (12336) & 33.12 (14640) & 33.17 (16944)\\
\hline

\end{tabular}
\end{center}
\vspace{-0.45cm}
\end{table}

\subsection{Towards Real-Time SR with FSRCNN}
Now we want to find a more concise FSRCNN network that could realize real-time SR while still keep good performance. First, we calculate how many parameters can meet the minimum requirement of real-time implementation (24 fps). As mentioned in the introduction, the speed of SRCNN to upsample an image to the size $760\times 760$ is 1.32 fps. The upscaling factor is 3, and SRCNN has 8032 parameters. Then according to Equation~\ref{eqn:computationSRCNN} and \ref{eqn:computationFSRCNN}, the desired FSRCNN network should have at most $8032\times 1.32/24\times 3^2\approx 3976$ parameters. To achieve this goal, we find an appropriate configuration --  FSRCNN (32,5,1) that contains 3937 parameters. With our C++ test code, the speed of FSRCNN (32,5,1) reaches 24.7 fps, satisfying the real-time requirement. Furthermore, the FSRCNN (32,5,1) even outperforms SRCNN (9-1-5)~\cite{Dong2014} (see Table~\ref{results91} and~\ref{results}). 

\subsection{Experiments for Different Upscaling Factors}
\label{sec:transfer}

\begin{figure}[t]\footnotesize
\centering
  \includegraphics[width=0.7\linewidth]{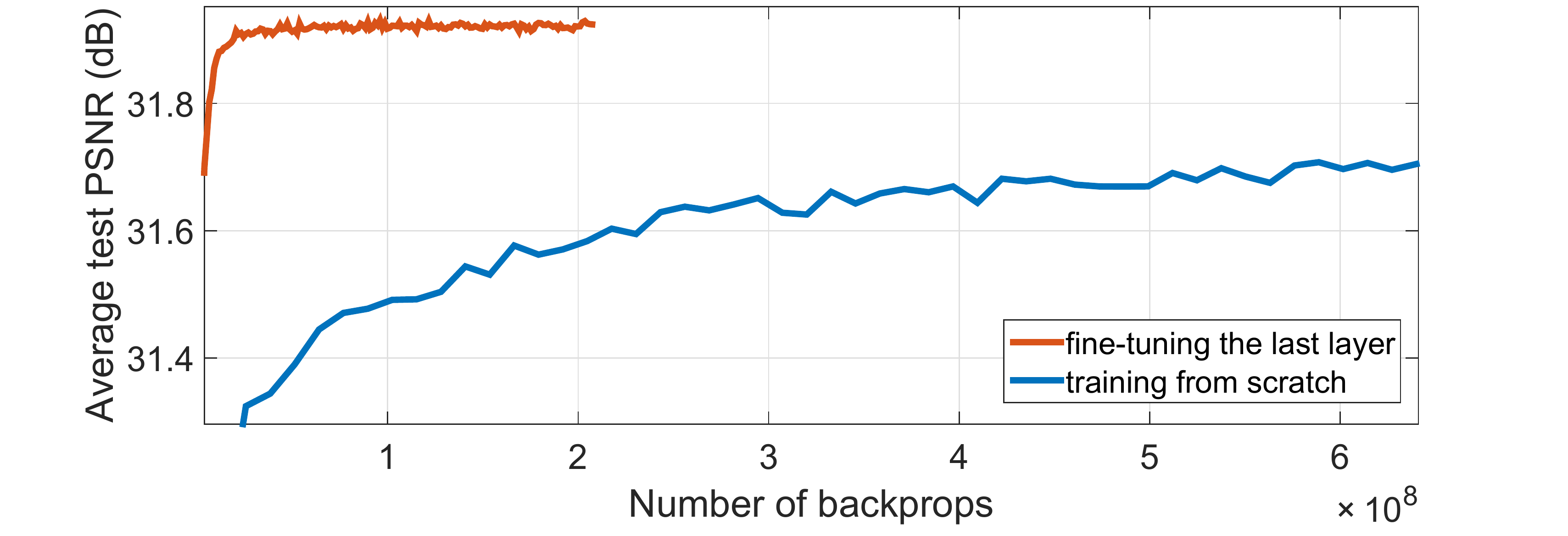}
\caption{Convergence curves for different training strategies.}
  \label{fig:upscalingfactor}
\end{figure}

Unlike existing methods~\cite{Dong2014,Dong2015} that need to train a network from scratch for a different scaling factor, the proposed FSRCNN enjoys the flexibility of learning and testing across upscaling factors through transferring the convolution filters (Sec.~\ref{subsec:across_factor}). We demonstrate this flexibility in this section.
We choose the FSRCNN (56,12,4) as the default network. As we have obtained a well-trained model under the upscaling factor 3 (in Section~\ref{sec:Investigation}), we then train the network for $\times$2 on the basis of that for $\times$3. To be specific, the parameters of all convolution filters in the well-trained model are transferred to the network of $\times$2. During training, we only fine-tune the deconvolution layer on the 91-image and General-100 datasets of $\times$2. For comparison, we train another network also for $\times$2 but from scratch. The convergence curves of these two networks are shown in Figure~\ref{fig:upscalingfactor}. Obviously, with the transferred parameters, the network converges very fast (only a few hours) with the same good performance as that training form scratch.
In the following experiments, we only train the networks from scratch for $\times$3, and fine-tune the corresponding deconvolution layers for $\times$2 and $\times$4.

\subsection{Comparison with State-of-the-Arts}
\label{subsec:sota}

\textbf{Compare using the same training set.} First, we compare our method with four state-of-the-art learning-based SR algorithms that rely on external databases, namely the super-resolution forest (SRF)~\cite{Schulter2015}, SRCNN~\cite{Dong2014}, SRCNN-Ex~\cite{Dong2015} and the sparse coding based network (SCN)~\cite{Wang2015}. The implementations of these methods are all based on their released source code. As they are written in different programming languages, the comparison of their test time may not be fair, but still reflects the main trend. To have a fair comparison on restoration quality, all models are trained on the augmented 91-image dataset, so the results are slightly different from that in the corresponding paper. We select two representative FSRCNN networks -- FSRCNN (short for FSRCNN (56,12,4)), and FSRCNN-s (short for FSRCNN (32,5,1)). The inference time is tested with the C++ implementation on an Intel i7 CPU 4.0 GHz.
The quantitative results (PSNR and test time) for different upscaling factors are listed in Table~\ref{results91}. We first look at the test time, which is the main focus of our work. The proposed FSRCNN is undoubtedly the fastest method that is at least $40$ times faster than SRCNN-Ex, SRF and SCN (with the upscaling factor 3),  while the fastest FSRCNN-s can achieve real-time performance ($>24$ fps) on almost all the test images. Moreover, the FSRCNN still outperforms the previous methods on the PSNR values especially for $\times$2 and $\times$3. We also notice that the FSRCNN achieves slightly lower PSNR than SCN on factor 4. This is mainly because that the SCN adopts two models of $\times$2 to upsample an image by $\times$4. We have also tried this strategy and achieved comparable results. However, as we pay more attention to speed, we still present the results of a single network.

\begin{figure*}[hp] \footnotesize
\begin{center}
\includegraphics[width=\linewidth]{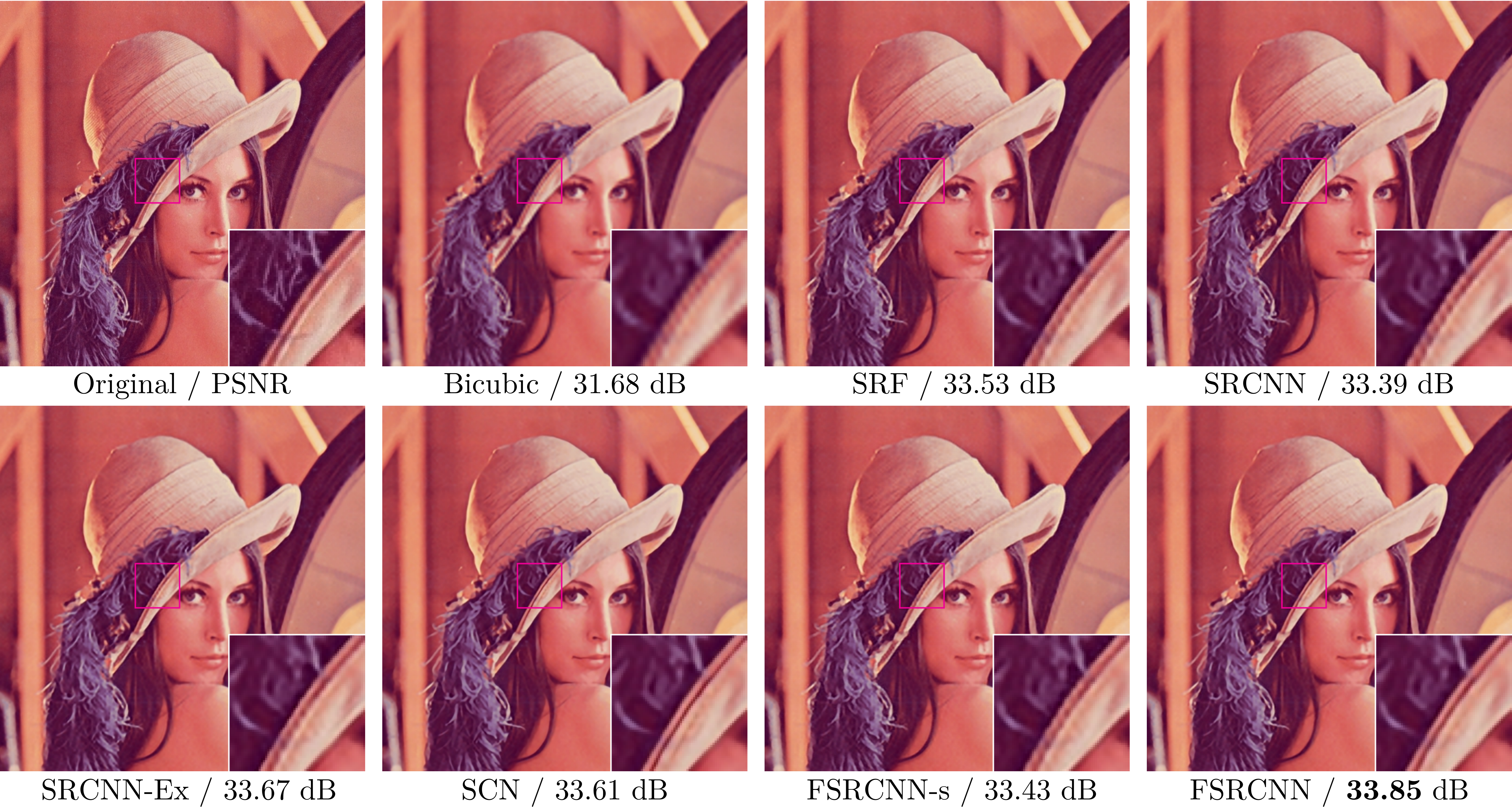}
\caption{The ``lenna" image from the Set14 dataset with an upscaling factor 3.}
\label{fig:result1}
\end{center}
\end{figure*}

\begin{figure*}[hp]\footnotesize
\begin{center}
\includegraphics[width=\linewidth]{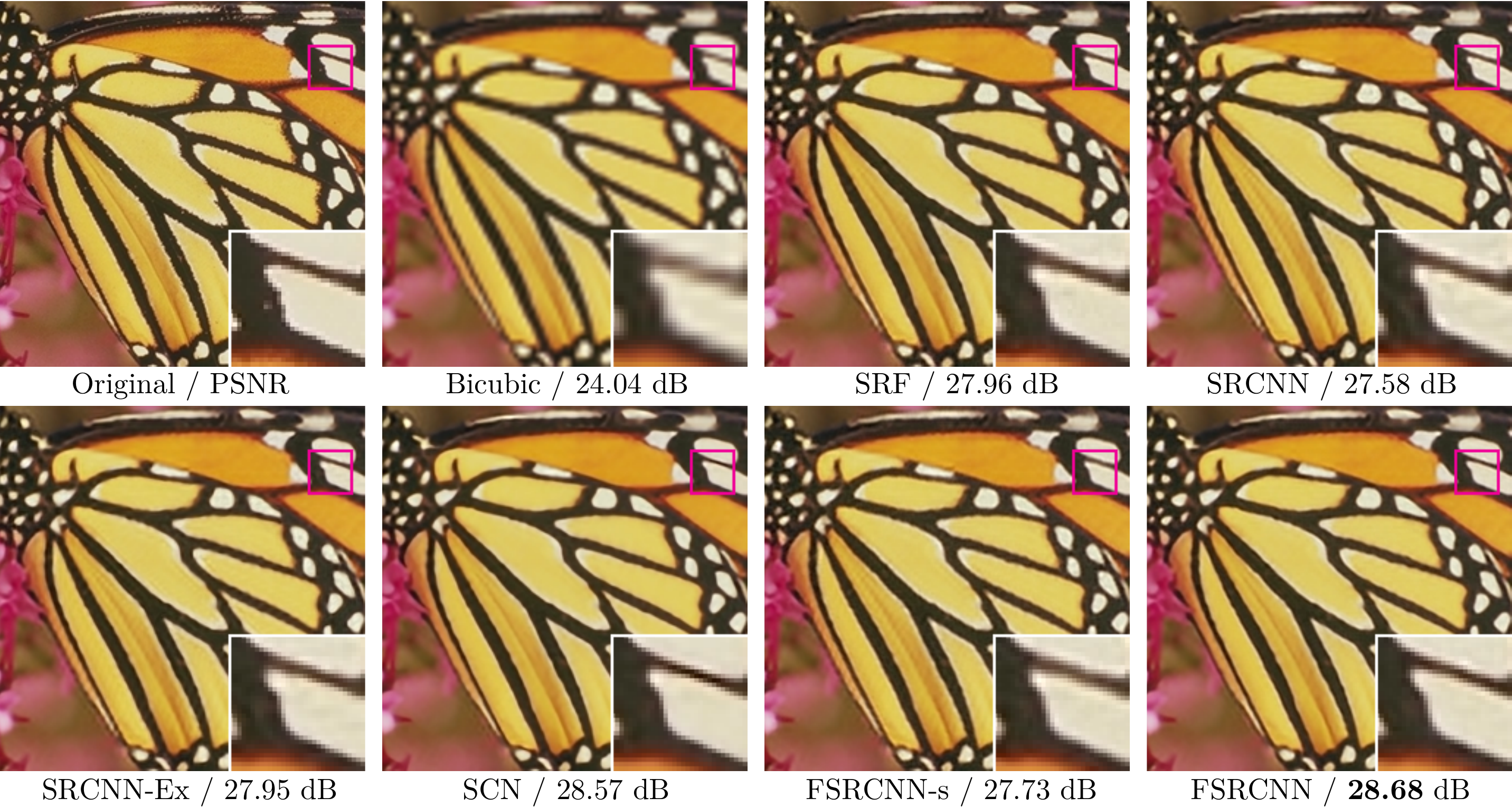}
\caption{The ``butterfly" image from the Set5 dataset with an upscaling factor 3.}
\label{fig:result2}
\end{center}
\end{figure*}

\noindent
\textbf{Compare using different training sets (following the literature).} To follow the literature, we also compare the best PSNR results that are reported in the corresponding paper, as shown in Table~\ref{results}. We also add another two competitive methods -- KK~\cite{Kim2010} and A+~\cite{Timofte2014} for comparison. Note that these results are obtained using different datasets, and our models are trained on the 91-image and General-100 datasets. From Table~\ref{results}, we can see that the proposed FSRCNN still outperforms other methods on most upscaling factors and datasets. We have also done comprehensive comparisons in terms of SSIM and IFC~\cite{sheikh2005information} in Table~\ref{result1} and~\ref{result2}, where we observe the same trend. The reconstructed images of FSRCNN (shown in Figure~\ref{fig:result1} and ~\ref{fig:result2}), more examples can be found on the project page) are sharper and clearer than other results. In another aspect, the restoration quality of small models (FSRCNN-s and SRCNN) is slightly worse than large models (SRCNN-Ex, SCN and FSRCNN). In Figure~\ref{fig:result1}, we could observe some "jaggies" or ringing effects in the results of FSRCNN-s and SRCNN.

\section{Conclusion}
While observing the limitations of current deep learning based SR models, we explore a more efficient network structure to achieve high running speed without the loss of restoration quality. We approach this goal by re-designing the SRCNN structure, and achieves a final acceleration of more than 40 times.
Extensive experiments suggest that the proposed method yields satisfactory SR performance, while superior in terms of run time. The proposed model can be adapted for real-time video SR, and motivate fast deep models for other low-level vision tasks.

\noindent
\textbf{Acknowledgment}. This work is partially supported by SenseTime Group Limited.

\begin{table*}\scriptsize
\caption{The results of PSNR (dB) and test time (sec) on three test datasets. All models are trained on the 91-image dataset. }\label{results91}
\begin{center}
\begin{tabular}{|c|c|c|c|c|c|c|c|c|c|c|c|c|c|c|c|}
\hline
 test & upscaling &  \multicolumn{2}{c|}{Bicubic} &  \multicolumn{2}{c|}{SRF~\cite{Schulter2015}} &  \multicolumn{2}{c|}{SRCNN~\cite{Dong2014}} &  \multicolumn{2}{c|}{SRCNN-Ex~\cite{Dong2015}} &\multicolumn{2}{c|}{SCN~\cite{Wang2015}} & \multicolumn{2}{c|}{FSRCNN-s} & \multicolumn{2}{c|}{FSRCNN}\\
 \cline{3-16}
 dataset & factor & PSNR&Time & PSNR&Time& PSNR&Time &PSNR&Time & PSNR&Time &PSNR&Time& PSNR&Time\\

\hline\hline
Set5 & 2 & 33.66 & -& 36.84 & 2.1 & 36.33 & 0.18 & 36.67 & 1.3 &36.76 & 0.94 & 36.53 & \textbf{0.024} & \textbf{36.94} &0.068 \\
Set14 & 2 & 30.23 & -& 32.46 & 3.9 & 32.15 & 0.39 & 32.35 & 2.8 & 32.48 & 1.7 & 32.22 & \textbf{0.061} & \textbf{32.54} &0.16 \\
BSD200 & 2 & 29.70 &- & 31.57 & 3.1 & 31.34 & 0.23 & 31.53 & 1.7 & 31.63 & 1.1 & 31.44 &\textbf{0.033}& \textbf{31.73} &0.098 \\
\hline\hline

Set5 & 3 & 30.39 & -& 32.73 & 1.7 & 32.45 & 0.18& 32.83 & 1.3 & 33.04 & 1.8 & 32.55 &\textbf{0.010}& \textbf{33.06} &0.027 \\
Set14 & 3 & 27.54 & -& 29.21 & 2.5 & 29.01 & 0.39& 29.26 & 2.8 &29.37 & 3.6 & 29.08 &\textbf{0.023}& \textbf{29.37} &0.061 \\
BSD200 & 3 & 27.26 &- & 28.40 & 2.0 & 28.27 & 0.23& 28.47 & 1.7 &28.54 & 2.4 & 28.32 &\textbf{0.013}& \textbf{28.55} &0.035 \\
\hline\hline

Set5 & 4 & 28.42 & -& 30.35 & 1.5 & 30.15 & 0.18 & 30.45 & 1.3& \textbf{30.82} & 1.2 & 30.04 & \textbf{0.0052} & 30.55 &0.015 \\
Set14 & 4 & 26.00 & -& 27.41 & 2.1 & 27.21 & 0.39 & 27.44 & 2.8& \textbf{27.62} & 2.3 & 27.12 & \textbf{0.0099} & 27.50 & 0.029 \\
BSD200 & 4 & 25.97 &- & 26.85 & 1.7 & 26.72 & 0.23 & 26.88 & 1.7 & \textbf{27.02} & 1.4 & 26.73 & \textbf{0.0072} & 26.92 &0.019 \\
\hline
\end{tabular}
\end{center}
\end{table*}

\begin{table*}\scriptsize
\caption{The results of PSNR (dB) on three test datasets. We present the best results reported in the corresponding paper. The proposed FSCNN and FSRCNN-s are trained on both 91-image and General-100 dataset. More comparisons with other methods on PSNR, SSIM and IFC~\cite{sheikh2005information} can be found in the supplementary file.}
\label{results}
\begin{center}
\begin{tabular}{|c|c|c|c|c|c|c|c|c|c|c|}
\hline
 test & upscaling &  {Bicubic} &  KK~\cite{Kim2010} &  A+~\cite{Timofte2014} & {SRF~\cite{Schulter2015}} &  {SRCNN~\cite{Dong2014}} &  {SRCNN-Ex~\cite{Dong2015}} &{SCN~\cite{Wang2015}} & {FSRCNN-s} & {FSRCNN}\\
 \cline{3-11}
 dataset & factor & PSNR & PSNR & PSNR  &PSNR  & PSNR  &PSNR & PSNR  &PSNR & PSNR \\

\hline\hline
Set5 & 2 & 33.66 & 36.20 & 36.55 & 36.89 & 36.34 & 36.66 & 36.93 & 36.58 & \textbf{37.00} \\
Set14 & 2 & 30.23 & 32.11 & 32.28 & 32.52 & 32.18 & 32.45 & 32.56 & 32.28 & \textbf{32.63} \\
BSD200 & 2 & 29.70 & 31.30 & 31.44 & 31.66 & 31.38 & 31.63 & 31.63 & 31.48 & \textbf{31.80} \\
\hline\hline

Set5 & 3 & 30.39 & 32.28 & 32.59 & 32.72 & 32.39 & 32.75 & 33.10 & 32.61 & \textbf{33.16} \\
Set14 & 3 & 27.54 & 28.94 & 29.13 & 29.23 & 29.00 & 29.30 & 29.41 & 29.13 & \textbf{29.43} \\
BSD200 & 3 & 27.26 & 28.19 & 28.36 & 28.45 & 28.28 & 28.48 & 28.54 & 28.32 & \textbf{28.60} \\
\hline\hline

Set5 & 4 & 28.42 & 30.03 & 30.28 & 30.35 & 30.09 & 30.49 & \textbf{30.86} & 30.11 & 30.71 \\
Set14 & 4 & 26.00 & 27.14 & 27.32 & 27.41 & 27.20 & 27.50 & \textbf{27.64} & 27.19 & 27.59 \\
BSD200 & 4 & 25.97 & 26.68 & 26.83 & 26.89 & 26.73 & 26.92 & \textbf{27.02} & 26.75 & 26.98 \\
\hline
\end{tabular}
\end{center}
\end{table*}

\begin{table*}[!h]\scriptsize
\caption{The results of PSNR (dB), SSIM and IFC~\cite{sheikh2005information} on the Set5~\cite{Sheikh2005}, Set14~\cite{Zeyde2012} and BSD200~\cite{martin2001database} datasets.}\label{result1}
\begin{center}
\begin{tabular}{|c|c|c|c|c|c|c|}
\hline
 test &  upscaling &  Bicubic &  KK~\cite{Kim2010} &  ANR~\cite{Timofte2013} & A+~\cite{Timofte2013} & SRF~\cite{Schulter2015} \\

 dataset & factor & PSNR/SSIM/IFC &  PSNR/SSIM/IFC &PSNR/SSIM/IFC & PSNR/SSIM/IFC & PSNR/SSIM/IFC\\

\hline\hline
Set5 & 2 & 33.66/0.9299/6.10 &  36.20/0.9511/6.87 & 35.83/0.9499/8.09 & 36.55/0.9544/8.48 & 36.87/0.9556/\textbf{8.63} \\
Set14 & 2 & 30.23/0.8687/6.09  & 32.11/0.9026/6.83 & 31.80/0.9004/7.81 & 32.28/0.9056/8.11 &32.51/0.9074/\textbf{8.22} \\
BSD200 & 2 & 29.70/0.8625/5.70 &  31.30/0.9000/6.26 & 31.02/0.8968/7.27 & 31.44/0.9031/7.49 & 31.65/0.9053/\textbf{7.60}\\
\hline\hline

Set5 & 3 & 30.39/0.9299/6.10  & 32.28/0.9033/4.14 & 31.92/0.8968/4.52 & 32.59/0.9088/4.84 & 32.71/0.9098/\textbf{4.90} \\
Set14 & 3 & 27.54/0.7736/3.41  & 28.94/0.8132/3.83 & 28.65/0.8093/4.23 & 29.13/0.8188/4.45 & 29.23/0.8206/\textbf{4.49} \\
BSD200 & 3 & 27.26/0.7638/3.19  &  28.19/0.8016/3.49 & 28.02/0.7981/3.91 & 28.36/0.8078/4.07 & 28.45/0.8095/4.11 \\
\hline\hline

Set5 & 4 & 28.42/0.8104/2.35  &  30.03/0.8541/2.81 & 29.69/0.8419/3.02 & 30.28/0.8603/\textbf{3.26} & 30.35/0.8600/3.26\\
Set14 & 4 & 26.00/0.7019/2.23 &  27.14/0.7419/2.57 & 26.85/0.7353/2.78 & 27.32/0.7471/2.74 & 27.41/0.7497/\textbf{2.94}\\
BSD200 & 4 & 25.97/0.6949/2.04  & 26.68/0.7282/2.22 & 26.56/0.7253/2.51 & 26.83/0.7359/2.62 & 26.89/0.7368/\textbf{2.62}\\
\hline
\end{tabular}
\end{center}
\end{table*}


\begin{table*}[!h]\scriptsize
\caption{The results of PSNR (dB), SSIM and IFC~\cite{sheikh2005information} on the Set5~\cite{Sheikh2005}, Set14~\cite{Zeyde2012} and BSD200~\cite{martin2001database} datasets.}\label{result2}
\begin{center}
\begin{tabular}{|c|c|c|c|c|c|c|}
\hline
 test &  upscaling &  SRCNN~\cite{Dong2014} & SRCNN-Ex~\cite{Dong2015} & SCN~\cite{Wang2015} & FSRCNN-s & FSRCNN \\

 dataset & factor &  PSNR/SSIM/IFC& PSNR/SSIM/IFC &PSNR/SSIM/IFC & PSNR/SSIM/IFC & PSNR/SSIM/IFC\\

\hline\hline
Set5 & 2 &  36.34/0.9521/7.54 & 36.66/0.9542/8.05 & 36.76/0.9545/7.32 & 36.58/0.9532/7.75 & \textbf{37.00}/\textbf{0.9558}/8.06 \\
Set14 & 2 & 32.18/0.9039/7.22 & 32.45/0.9067/7.76 & 32.48/0.9067/7.00 & 32.28/0.9052/7.47 & \textbf{32.63}/\textbf{0.9088}/7.71\\
BSD200 & 2 &  31.38/0.9287/6.80 & 31.63/0.9044/7.26 & 31.63/0.9048/6.45 & 31.48/0.9027/7.01 & \textbf{31.80}/\textbf{0.9074}/7.25\\
\hline\hline

Set5 & 3 &  32.39/0.9033/4.25 & 32.75/0.9090/4.58 & 33.04/0.9136/4.37 & 32.54/0.9055/4.56 & \textbf{33.16}/\textbf{0.9140}/4.88\\
Set14 & 3 &  29.00/0.8145/3.96 & 29.30/0.8215/4.26 & 29.37/0.8226/3.99 & 29.08/0.8167/4.24 & \textbf{29.43}/\textbf{0.8242}/4.47 \\
BSD200 & 3 & 28.28/0.8038/3.67 & 28.48/0.8102/3.92 & 28.54/0.8119/3.59 & 28.32/0.8058/3.96 & \textbf{28.60}/\textbf{0.8137}/\textbf{4.11} \\
\hline\hline

Set5 & 4 &  30.09/0.8530/2.86 & 30.49/0.8628/3.01 & \textbf{30.82}/\textbf{0.8728}/3.07 & 30.11/0.8499/2.76 & 30.71/0.8657/3.01\\
Set14 & 4 &  27.20/0.7413/2.60 & 27.50/0.7513/2.74 & \textbf{27.62}/\textbf{0.7571}/2.71 & 27.19/0.7423/2.55 & 27.59/0.7535/2.70 \\
BSD200 & 4 & 26.73/0.7291/2.37 & 26.92/0.7376/2.46 & \textbf{27.02}/\textbf{0.7434}/2.38 & 26.75/0.7312/2.32 & 26.98/0.7398/2.41\\
\hline
\end{tabular}
\end{center}
\end{table*}

\clearpage

\bibliographystyle{splncs}
\bibliography{short,cnn_sr}
\end{document}